\documentclass{article} 
\usepackage{iclr2020_conference,times}

\usepackage{amsmath,amsfonts,bm}

\def\eqref#1{equation~\ref{#1}}

\def\1{\bm{1}}

\DeclareMathAlphabet{\mathsfit}{\encodingdefault}{\sfdefault}{m}{sl}
\SetMathAlphabet{\mathsfit}{bold}{\encodingdefault}{\sfdefault}{bx}{n}

\usepackage{hyperref}
\usepackage{url}
\usepackage{graphicx}

\title{RTOP: A Conceptual and Computational Framework for General Intelligence}

\author{Shilpesh Garg \\
Akribia Technologies, India \\
\texttt{shilpesh@akribiatech.com} \\
}

\iclrfinalcopy
\begin{document}

\maketitle

\begin{abstract}
A novel general intelligence model is proposed with three types of learning. A unified sequence of the foreground percept trace and the command trace translates into direct and time-hop observation paths to form the basis of Raw learning. Raw learning includes the formation of image-image associations, which lead to the perception of temporal and spatial relationships among objects and object parts; and the formation of image-audio associations, which serve as the building blocks of language. Offline identification of similar segments in the observation paths and their subsequent reduction into a common segment through merging of memory nodes leads to Generalized learning. Generalization includes the formation of interpolated sensory nodes for robust and generic matching, the formation of sensory properties nodes for specific matching and superimposition, and the formation of group nodes for simpler logic pathways. Online superimposition of memory nodes across multiple predictions, primarily the superimposition of images on the internal projection canvas, gives rise to Innovative learning and thought. The learning of actions happens the same way as raw learning while the action determination happens through the utility model built into the raw learnings, the utility function being the pleasure and pain of the physical senses.
\end{abstract}

\section{Introduction}
There has been a lot of research in AI in the last many decades to understand general-intelligence and be able to replicate it in machines. \cite{goertzel2014artificial} discusses various approaches adopted by different research communities over a period of time. The different theories and models have provided us great insights into human cognition and action; however, a model that can describe and demonstrate human-like intelligence in a wide variety of domain is still elusive. 

The approaches adopted so far can be broadly classified into two main categories: Symbolic and Subsymbolic. While the functioning of the symbolic implementations have been more investigable, explainable and tunable, the symbols themselves were usually very different from the real-world data and the logic often became too convoluted and abstract. Even with the flexibility of statistics, these systems have proved brittle and often required numerous patches when presented with the real-world data. On the other hand, the subsymbolic systems of today, especially those employing artificial neural networks, have shown great progress and adaptiveness with the real-world data. The deep-learning networks \citep{lecun2015deep} have achieved near human levels of image recognition on limited image datasets and have demonstrated very good speech recognition as well. When architected as a generative adversarial network \citep{goodfellow2014generative}, such models have shown the ability to even synthesize very realistic representations of the world. However, these approaches usually require huge amounts of labelled data for training, require a great deal of computational resources, are usually domain specific, and are often very opaque \citep{marcus2018deep}.

We believe that purely subsymbolic approaches may not be sufficient to achieve human like intelligence. If the neural network models start taking as inputs, the variety of sensory and action data of the sensorimotor machine they are controlling, they might eventually start exhibiting some sort of general intelligence. However, in the absence of a symbolic component, they may not be able to mimic the human thought process, which we believe is purely symbolic. Also, our lack of ability to interpret the subsymbols and their associations, would cause the training and the improvement of such systems to keep getting more and more difficult. In addition, we believe that a scientific theory ought to provide insights as much as the accuracy of results and prediction.

To overcome these limitations, we propose a novel approach for building general intelligence. Our approach uses symbols too, however, the symbols are the real world data and the intelligence operates through a few consistent algorithms making use of this data. The knowledge base is built using perceived images, audio and other sensory data as modal symbols \citep{barsalou1999perceptual}. Our model, however, delves deeper into the formation, the transformation and the usage of the knowledge base; we term this model as RTOP which stands for “Reduction of temporal observations and predictions”. The main idea behind this model is that learning is of three kinds: Raw, Generalized and Innovative. The Raw or fundamental learning happen through the coming together of the sensory data, the actions and pleasure-pain to form a common consistent internal representation of all kinds of observations. The knowledge base thus formed is continuously looked up to match the ongoing observations, to build predictions and to determine the best actions, like a traditional utility based agent \citep{russell2016artificial}. Generalized learning happens through regular offline identification of similarities in the observations and their subsequent reduction into common generic observations. Generalization makes the knowledge base more robust to matching and extracts common associations, rules and sensory properties from raw observations. Innovative learning involves online or conscious merging of symbols across different predictions to give rise to novel internal representations that may well be significantly different from any past observations. These superimpositions form the higher level of intelligence, like story-understanding and the thought process. 

The three learning mechanisms form the conceptual framwork of our proposed model. In addition, we propose a computational framework that can help to build an implementation of the model. The framework suggests how memory is organized, how matches are detected, how probabilities of occurrence are arrived at, and how happiness is calculated to determine action. The paper first discusses the building blocks of the framework like sensory memory, action memory, pleasure-pain, attention, and observation trees. It then discusses the details of the three types of learnings in terms of the building blocks. It also discusses a partial implementation of the framework, which we call the RTOP Agent program, and which we use to evaluate some of the proposed concepts.

\section{Raw Learning}

\subsection{Building blocks of learning}

\subsubsection{Memory Nodes}
The smallest unit of memory for an agent is a memory-node. Every sense and every action have a corresponding type of memory node. A vision memory node contains a perceived image at an instant while an audio node contains the perceived changes in pressure over time for a small duration. Similarly, different types of nodes exist for smell, touch, taste. In addition to the memory of the sensory inputs of the external environment, the model proposes memory of the sensory inputs from the internal components. For a human like agent, it would include hunger, muscular pain, sexual arousal, mental fatigue and so on. The internal senses can generally take only few different states, and through these, indicate to the agent the overall wellbeing of the system. A positive indication we term as pleasure and a negative indication we term as pain. In the proposed model, pleasure and pain are scalar values, perceived at all times and stored along with all other raw sensory input storage.

Action memory usually pertains to speech or some motor movement as a way to interact with the environment. However, some actions play an important role in the perception itself; for example, movement of visual focus, change in sharpness or brightness of the visual input, enhancing certain frequencies of the audio input and change in attention from audio to visual input or vice versa. Another special action that plays an important role in learning and is internal to the agent processing is the Superimpose action which the agent uses to merge the representations of the recalled memory nodes.

The notation used for memory nodes in this paper is \texttt{<Type>.<Identifier>}. For example, IMG.10 indicates image node numbered 10; AUD.22 indicates audio node numbered 22; HNG.HIGH indicates hunger node for high hunger; ATT.IMG indicates action to change attention to visual input; IFA.x,y,z indicates image focus change action for moving focus right by x pixels, down by y pixels and increasing the focus area, that is, zoom, by z pixels; SPK.p-A indicates speech action with phone sequence p, A. 

Memory nodes can point to an unlimited number of other memory nodes. The interlinking of nodes gives rise to observation paths and knowledge graphs that serve to store experiences over a duration of time. Organizing overall memory in the form of interconnected memory nodes simplifies the architecture, simplifies the reproduction of memory for agent's internal projection and simplifies the inspection of the system for the researcher.

\subsubsection{Indexing and Matching of nodes}
To be able to predict the future based on the current percept sequence, the agent needs to look up the historical knowledge base and quickly identify the closely resembling memory nodes and observation paths. For this reason, matching of nodes is an important functionality in the model. In the agent program this is done by creating indexes on a few summary attributes for each type of memory. For example, for image nodes, the mean-lightness is one of the indexed attributes. Searching with a leeway on the indices provides a list of candidate nodes. A more extensive comparison is then performed on the detail attributes of the candidate nodes to find the matches. In the agent program, for image nodes, every pixel of a candidate node is compared with the corresponding pixel of the node being searched to determine a match.

\subsubsection{Attention}
In the proposed model, not all data that is perceived is saved to the memory; it depends upon an important variable: the agent’s Attention. As discussed by Helgi \citep{helgason2012attention}, attention exists to carefully select which information will be processed and which will be ignored to make an economical use of the available computing resources of the agent.

The agent is capable of processing multiple sensory inputs at the same time, however, there is only a maximum of one input that can hold the agent’s attention at any given instant of time. The specific sensory input processing that the agent is attending to goes through Foreground Processing while all the other processes go through Background Processing. The agent may alternatively be busy in the thought process, that is, forming internal projections, in which case, none of the sensory inputs would hold the attention. This means that when the agent is watching, it is not listening, it is just hearing in the background.

In both foreground and background processing, the input is captured, a matching attempt is made and if a match is found, a prediction is attempted. However, in the case of foreground processing, the captured inputs get saved to the memory, while in the case of background processing, the captured inputs are discarded and not stored. While foreground processing leads to new learning and innovation, the main purpose of the background processing is to use the existing learning in a quick way to take routine actions and to detect any major changes or opportunities or threat. If it detects such an event, it requests the agent’s attention.

The observations that an agent has learnt well through repeated experience, become fairly easily retrievable and predictable and thus start requiring less or no attention and less processing power of the agent. Such observations become eligible for background processing. As the agent matures, there are fewer interactions that require foreground processing, causing the attention to increasingly shift to thought and problem solving. 

\subsection{Observation trace}
That the agent can yield attention to only a single process means all foreground sensory input and thought can be serialized to a single queue. Observation trace is the sequence of all foreground inputs and all actions impacting the foreground processing. If the attention is visual input, foreground actions would usually be - moving the visual focus or changing attention from visual focus to somewhere else. 

Suppose the agent first observes an image, then zooms the focus, then observes another two images, then changes attention to audio, then hears a small audio of duration two nodes. The observation trace is:\\
\begin{footnotesize}
\verb+IMG.1 -> IFA.ZOOMIN -> IMG.2 -> IMG.3 -> ATT.AUD -> AUD.1 -> AUD.2+
\end{footnotesize}

Representing perception and action in such an observation trace is central and very significant to the Raw Learning hypothesis. Inclusion of actions in observations provides many benefits over a simple percept trace. The above sequence inherently contains (a) temporal relationships, like that between IMG.2 and IMG.3 or between AUD.1 and AUD.2 (b) spatial relationship between IMG.1 and IMG.2 and (c) loose relationship between the image nodes and the audio nodes. This paves way for learning different aspects of the environment in a consistent way. The strong relationship between action and perception has been explored in detail by Alva Noe \citep{noe2004action}. This relationship also implies that training such an agent cannot be as simple as feeding arbitrary pictures or audio-visual streams, but will need interaction, either real or simulated.

A segment of the observation trace we term as an observation path. The sequence of nodes in a path is in the same order in which they are perceived in time, so we can call it a temporal observation path. There would be a similar trace for background processing as well, however that trace won’t have any attention change nodes, and the nodes that are there won’t get saved to the agent’s memory. In the agent program, observation trace is cleared after every offline processing of generalization. 

\subsection{Storage and indexing of observation paths}
To be able to organize and index well the multiple experiences contained within one long observation trace, the agent would need to break it down into multiple smaller sequences called observation paths. One small sequence can represent one small experience, for example, observing an object move or listening to a small piece of a song. In the agent program, one observation path is typically 15 to 20 memory nodes long and is referenced by only the first node of the sequence. Hence the first memory node of a path serves as the index for retrieving the whole path. Now, the first node of a given observation path may also be the first node of many other observation paths. This would lead to the development of an observation tree associated with the indexing nodes as below.\\
\begin{footnotesize}
\verb+IMG.1---|--(0.7)--> IMG.2 -----(1.0)--> IMG.3 --(1)--> IMG.4+
\verb+        |--(0.3)--> IMG.5 --|--(0.7)--> IMG.6 --(1)--> IMG.7+
\verb+                            |--(0.3)--> IMG.8 --(1)--> IMG.9	 +
\end{footnotesize}\\
This is the observation tree of IMG.1. The numbers in brackets represent the probability of occurrence based on the number of past occurrences observed; similar to Bayesian inference. Longer observation paths would mean the agent can predict better and longer, however, they would also increase the complexity of the tree and increased time and resource requirements in organizing and retrieving the data, so a balance is required. 

For a small experience to be retrievable on just its first memory node reduces the recall and prediction capability of the agent. To avoid this, the breakdown of observation trace can be done using a moving window.

While the Direct Paths discussed above provide the agent the ability to predict immediate future, like words, sentences and object movements, the agent also needs the ability to predict medium-term and long-term future in a quick way to identify the most beneficial actions. For this purpose, the model suggests the concept of Jump Paths where the observation paths are created by successively jumping few nodes of the observation trace. A Jump path may look like: \begin{footnotesize}\verb+IMG.1->JMP.5->IMG.7->JMP.5->IMG.13+\end{footnotesize} where JMP.5 indicates a jump of five nodes. The raw memory nodes of a jump path can still serve as forks for the agent to build the intermediate story by peeping inside the observation trees of these nodes. Jump paths are like highways on the memory route network.

\subsection{Relationship Formation}

\subsubsection{Image-Image temporal relationship}
\begin{figure}[h]
\begin{center}
\includegraphics[width=0.6\linewidth]{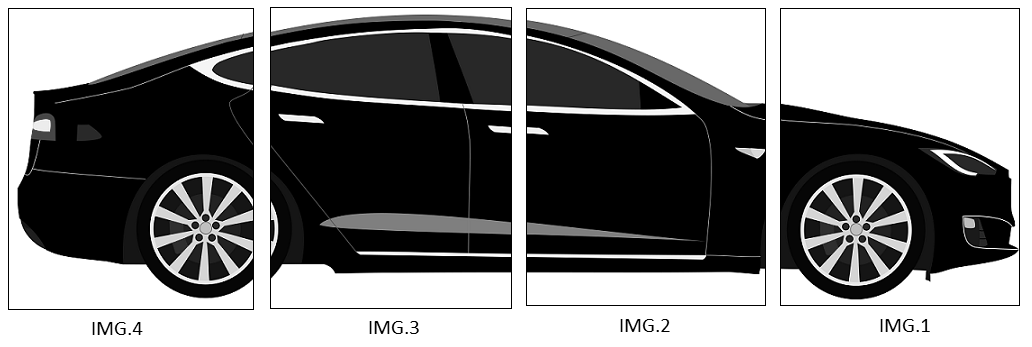}
\end{center}
\caption{Moving object}
\label{fig:movingobject}
\end{figure}

Suppose, the agent’s visual focus is stationary, and a car is moving from left to right before it. If we represent the car with four image nodes as shown in figure \ref{fig:movingobject}, the observe path is: \begin{footnotesize}\verb+IMG.1->IMG.2->IMG.3->IMG.4+\end{footnotesize} This sequence would get saved in the agent’s memory against memory node IMG.1. In future, if the same sequence happens again, that is, if the agent observes \begin{footnotesize}\verb+IMG.1->IMG.2+\end{footnotesize}, it would be able to look up the sequence from memory and predict that \begin{footnotesize}\verb+IMG.3->IMG.4+\end{footnotesize} are soon to follow. The agent is able to predict the order in which the upcoming images are expected. This means that when drawing the predictions on the internal canvas, the agent can even project a video using the predicted images. 
 
The observation trees are maintained separately for each indexing node, rather than creating a common global observation tree. A global observation tree would suffer from significant loss of information. In the car example just discussed, if the same car now reverses from right to left, the observation path would be \begin{footnotesize}\verb+IMG.4->IMG.3->IMG.2->IMG.1+\end{footnotesize}. Now, keeping these two observation paths separate, as part of nodes IMG.1 and IMG.4 respectively, enables the agent to predict more accurately when either \begin{footnotesize}\verb+IMG.1->IMG.2+\end{footnotesize} is observed, or \begin{footnotesize}\verb+IMG.4->IMG.3+\end{footnotesize} is observed; which is not the case if we keep a global graph as below\\
\begin{footnotesize}
\verb+                  |-----> IMG.1  +\\
\verb+IMG.1 --> IMG.2 --|--|--> IMG.3 --|--> IMG.4+\\
\verb+          IMG.4 -----|            |--> IMG.2+
\end{footnotesize}

\subsubsection{Image-Image spatial relationship}
Suppose, a car is parked stationary before the agent’s camera. Suppose image IMG.1 represents the car and IMG.2 represents a wheel of the car. Suppose the agent first sees the car, then zooms-in the camera focus by 100 pixels, moves the focus down by 50pixels and moves the focus left by 40 pixels, and now sees the wheel. The agent would observe sequence of nodes as 
\begin{footnotesize}\verb+IMG.1->IFA.40,50,100->IMG.2+\end{footnotesize}
Now suppose the agent moves the focus back to the car, the observation path would be 
\begin{footnotesize}\verb+IMG.2->IFA.-40,-50,-100->IMG.1+\end{footnotesize}
The first sequence gets saved against node IMG.1 while the second one against IMG.2. In future, if the agent sees IMG.1 it would expect that if it moves the focus by 40,50,100, it would see IMG.2. Similarly, if the agent sees IMG.2 in future, it would expect that if it moves the focus by -40,-50,-100, it would see IMG.2. 

Spatial relationships are important in how the agent perceives the real-world objects. An object to an agent is just a group of images related together through temporal or spatial means or through sounds or words. Identifying a real-world object may involve matching just a single image node or matching multiple nodes of an observation path or matching multiple nodes across multiple observation paths. 
Suppose the wheel of car and the wheel or a motorcycle look very similar and are represented by IMG.1, the image of the car is represented by IMG.2 and that of the motorcycle is represented by IMG.3. The agent knows, that by simply looking at the wheel it cannot determine whether it is a car or a motorcycle. It needs to shift visual focus, IFA.-40,-50,-100 in the tree below, to do so\\
\begin{footnotesize}
\verb+IMG.1->IFA.-40,-50,-100 --|--(0.5)->IMG.2->ATT.AUD->AUD.CAR+\\
\verb+                          |--(0.5)->IMG.3->ATT.AUD->AUD.MOTORCYCLE+
\end{footnotesize}\\
If there are multiple image parts required to be matched to arrive at the object configuration, the agent would need to execute the \verb+matching->prediction->focus_change_action+ steps multiple times. The proposed model is in contrast to the hierarchical models. Here the images of parts of an object, the object itself and the surroundings of an object are all stored in consistent nodes and linked through the same relationship tree, and not through any higher-level subsymbolic memory nodes containing a summary of the lower layer, like in HTM \citep{hawkins2006hierarchical}. 

If the agent is given a task of identifying one or more objects of different sizes scattered in a picture, the agent would first retrieve matches based on the image of the overall picture. Then, if the retrieved paths indicate to focus on particular areas of the picture, the agent would focus accordingly and perform further matching to detect any objects. Otherwise, it would scan through the picture with varying focus sizes and varying focus centres. This is similar to how humans analyse pictures. When a person looks at a page of a book, he does not know what is written on it, but he knows that he needs to zoom and focus on the top left to expect an image of a word and then to successively move the focus right to see more words. In this approach, the image matching itself is a lesser challenge and may not require techniques like feature detection, edge detection or segmentation. However, it adds complexity around scanning the visual area and the multiple steps of matching.

\begin{figure}[h]
\begin{center}
\includegraphics[width=0.5\linewidth]{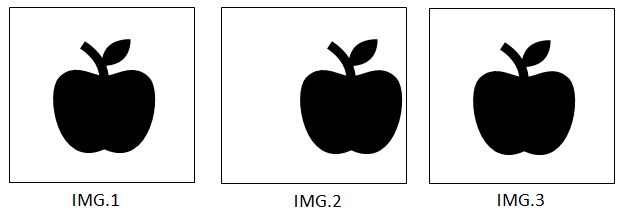}
\end{center}
\caption{Object shift}
\label{fig:objectshift}
\end{figure}

Good matching of images is important in good retrieval of observation paths. In the agent program, every pixel of a candidate match is compared with the corresponding pixel of the node being searched to determine a match. Using such an approach for image matching means that even small variations like addition of noise, change of background, horizontal or vertical translation of an image may all lead to candidate images being classified as mismatches. In the proposed model, the agent would develop the ability to match images despite these differences over time through (a) the reduction, merging and masking process that will be discussed in the generalization section (b) learned actions to move the visual focus or to enhance the visual input. To illustrate the latter, suppose the agent first looks at apple image IMG.1 as shown in figure \ref{fig:objectshift}. Now if the apple is moved slightly to the right, the image captured is IMG.2. Pixelwise comparison between IMG.1 and IMG.2 yields a mismatch, so the two images are stored separately. Now if the agent happens to move the visual focus to the right by a few pixels (say 20), IMG.3 would be captured. IMG.3 being very similar pixelwise to IMG.1 would be considered a match. The observation path is:
\begin{footnotesize}
\verb+IMG.1 --> IMG.2 --> IFA.0,20,0 --> IMG.1 +
\end{footnotesize}\\
If IMG.1 is something that gives some pleasure to the agent, then upon displacement of the object, it would learn to take the focus move action to redetect IMG.1. Such successive image focus change action upon continual displacement of an object would give rise to the tracking focus ability in the agent. 

\subsubsection{Audio relationships}
Suppose the agent is looking at a car represented by IMG.CAR. Suppose now someone speaks the word “Car” and it causes the agent to shift attention to the audio processing. If the audio node to represent the word car is AUD.CAR, the observation path is:\\
\begin{footnotesize}
\verb+IMG.CAR -> ATT.AUD -> AUD.CAR+\\
\end{footnotesize}
Suppose the agent hears the word “Car” and then a car appears in front of its camera causing the agent to shift attention to visual processing. In this case the observation is:\\
\begin{footnotesize}
\verb+AUD.CAR -> ATT.IMG -> IMG.CAR +
\end{footnotesize}\\
The two observation paths get stored against IMG.CAR and AUD.CAR respectively. If in future, the agent sees the same car, the agent can predict the audio associated with it and can play the audio internally. If it hears the word Car, it can predict the image associated with it and can project it internally. 

The word “Car” may be uttered in a variety of ways at different times due to different timbre, different frequencies, different loudness, different speed, different stress, and so on. In the proposed model, the ability to match audio nodes despite these differences would evolve in the agent over time through (a) the reduction, merging and masking process that will be discussed in the generalization section (b) learned actions to enhance or suppress frequencies or to expand or compress the waveform in time. The latter one is similar to the tracking focus development that we discussed in the previous section. Even after the development of these abilities, there may still be multiple audio nodes for the same word, many of which may point to the same images in their observation trees. 

In case the agent is attending to audio for a longer duration, for example listening to a song, the audio would be stored as a temporal sequence of multiple audio nodes, the same way a moving car is stored as a temporal sequence of image nodes. 

\subsection{Prediction}
Prediction is finding the sequences of memory nodes expected to happen in future. In the proposed model, these are represented in the same form as an observation tree and are called future-trees. A future tree contains expected probabilities and expected pleasure-pain changes on each connection just like an observation tree. 

In the agent program, future-trees are built by simply looking up the ongoing observation in the knowledge base. The ongoing observation itself though is very dynamic with new inputs being captured in quick succession. The newly captured memory nodes usually do not invoke fresh build of future-trees as long they satisfy or conform to the expected memory nodes of the ongoing future-trees and unless the probabilities or lengths of the remaining parts of the ongoing future trees become too small.

Prediction happens for the observations in the foreground processing as well as for those in the background processing. A human, while driving a car, may be engaged in some thought, but at the same time is processing the audio-visual traffic data and making various predictions about its environment. Both short-term and long-term predictions can happen at the same time and both can invoke corresponding actions. Long term predictions involve jump nodes as discussed earlier. This is in contrast to a hierarchical agent system where controllers are operating at different levels for different levels of percepts and commands. In the proposed model, there is a single controller acting through different levels of predictions.

\subsection{Action Learning}
In the initial stages of an agent’s life, actions are generated either randomly or as hard-coded responses to big changes in the environment. As the agent starts observing the correlations between these actions and the corresponding changes in the environment, it learns those actions and they become learned actions. Actions include not just physical actions but mental actions like changing attention, changing image focus and applying audio filters. The more the agent observes, the more it would learn about its actions and their impact, and the smaller would become the set of actions that are still not fully explored or understood by the agent. Some of the mental actions, like shifting attention from visual to audio due to a sudden loud noise, happen as hard-wired reflexes in the agent.

Once actions are learnt, in a given situation, agent may notice that there are several possible actions that it can take or that it has taken in the past. In such a situation, it needs to determine which particular action to actually take. This determination is done by evaluating expected net pleasure over the various available paths and then taking the action which leads to the path with the maximum net pleasure.

\subsubsection{Happiness calculation}
In the proposed model, a future tree contains pleasure-pain changes on each connection or branch, in addition to the probabilities of occurrence. An observation tree with the changes in pleasure/pain would typically looks like:\\
\begin{footnotesize}
\verb+IMG.1---|--(0.7,0.0)-->IMG.2-----(1.0,0.0)-->IMG.3--(1,0.0)--->IMG.1 +\\
\verb+        |--(0.3,0.5)-->IMG.4--|--(0.7,1.0)-->IMG.5--(1,-0.5)-->IMG.3 +\\
\verb+                              |--(0.3,0.0)-->IMG.9--(1,0.0)--->IMG.7	 +	
\end{footnotesize}\\
where numbers in the brackets represent the correlation and the \(\Delta P\textsubscript{net}\) respectively. \(\Delta P\textsubscript{net}\) refers to a change in net pleasure or pain, where, net pleasure or pain itself is a sum of pleasure or pain values across all senses. Pleasure or pain \(P\textsubscript{k}\) could be due to hunger, touch, loudness of audio, muscle pain, and so on. Each such type of pleasure or pain could be different in nature, and it is important for the agent to keep the memory of the different types separately to be able to predict better. However, a common representation of these disparate pleasure pains and an aggregate value across the different types is also important to be able to compare one type against the other and be able to take an action by prioritizing one over the other in the future trees. One simple representation of such a net value could be\\
\verb+    +\( P\textsubscript{net} = c\textsubscript{1}.P\textsubscript{Hunger} + c\textsubscript{2}.P\textsubscript{Touch} + c\textsubscript{3}.P\textsubscript{MusclePain} + …\)\\
with some constant coefficients \(c\textsubscript{k}\) for each type of \(P\textsubscript{k}\). \(P\textsubscript{net}\) at any point of time represents the net sensory or physical pleasure or pain, whereas \(\Delta P\textsubscript{net}\) represents the change in the \(P\textsubscript{net}\) over a small duration. The \(\Delta P\textsubscript{net}\) of an observation path represents the change in \(P\textsubscript{net}\) over the path. The \(\Delta P\textsubscript{net}\) of a future tree represents the expected change in pleasure pain over the future-tree prediction. The \(\Delta P\textsubscript{net}\) of a future-tree with N nodes, can be calculated as\\
\verb+    +\( \Delta P\textsubscript{net\textbar FutureTree} = \rho\textsubscript{1}.\Delta P\textsubscript{net\textbar node1}  + \rho\textsubscript{2}.\Delta P\textsubscript{net\textbar node2} + …\)\\
where \(\rho\textsubscript{k}\) is the net-probability of the \(k\textsuperscript{th}\) node, which itself can be calculated by multiplying the probabilities of all the nodes in the observation path prior to the \(k\textsuperscript{th}\) node.
For future-tree of IMG.1 shown at the top of this section: \\
\verb+   +\(\Delta P\textsubscript{net} = (0.3*0.5) + (0.3*0.7*1.0) + (0.3*0.7*1.0*-0.5) = 0.15 + 0.21 - 0.105 = 0.255\)\\
The \(\Delta P\textsubscript{net}\) of all future-trees at any given instant is what we term as the Happiness at that instant. A positive value represents joy while a negative value represents sorrow. While pleasure and pain are purely physical attributes associated with the senses, joy and sorrow are attributes of the online belief state of the agent.

\subsubsection{Action determination}

Happiness is the determiner of the learned actions. The agent would take that particular action which promises the maximum happiness. 

Suppose the below observation tree is associated with the image of a car. The tree indicates that in the past, when an image of pen was shown, 40\% of the times the agent spoke the word “Pen”, 40\% of the times it spoke “Cat” and 20\% of the times it spoke “Car”. In first two cases, the speaking was followed by hearing the word “Wrong” with no change in pleasure-pain, whereas in the last case, the speaking was followed by hearing the word “Right” with a positive change in pleasure-pain. This would cause the agent to select the third path in order to maximize the happiness. It would speak “Car” and then change attention to audio\\
\begin{footnotesize}
\verb+IMG.CAR-|-(0.4,0)->SPK.PEN-(1,0)->ATT.AUD-(1,0)->AUD.PEN-(1,0)->AUD.WRONG+\\
\verb+        |-(0.4,0)->SPK.CAT-(1,0)->ATT.AUD-(1,0)->AUD.CAT-(1,0)->AUD.WRONG+\\
\verb+        |-(0.2,0)->SPK.CAR-(1,0)->ATT.AUD-(1,0)->AUD.CAR-(1,1)->AUD.RIGHT+\\
\end{footnotesize}\\
Thus, an agent can be made to learn actions better by providing appropriate rewards and penalty 

\subsection{Hypothesis Evaluation}

We evaluate the hypotheses of raw learning through an implementation we term as the RTOP agent program. A few simple experiments were conducted to establish relationship formation and retrieval for prediction. The experiments are discussed after a brief discussion about the agent program itself. 

\subsubsection{RTOP Agent Program}

The program implements some of the concepts discussed in this paper. It captures images and audio from the computer camera and the computer microphone respectively. It simulates hunger and comfort as internal senses. An interacting user can increase or decrease the comfort through a user-interface. Foreground processing captures the inputs and actions and saves them to the appropriate memory nodes and creates temporal observation paths and observation trees. Background processing only detects any substantial changes in the audio or visual input and accordingly moves the agent’s attention. The actions built in the program are attention change, image focus change and speech action, which are mostly generated randomly. The program makes predictions by looking up the memory node of the ongoing observation in the knowledge base and retrieving the relevant observation trees along with the connection probabilities and the expected changes in pleasure pain. The program takes learned actions based on happiness. It also implements some of the principles of generalized learning which will be discussed later. Further details about the program are mentioned in appendix \ref{agentprogram} and the architecture and flow is shown in \ref{rtoparch}. The raw learning experiments are discussed below. 

\subsubsection{Experiment: Relationship learning}

The objective behind this experiment was to establish that temporal observation paths can indeed serve as the basis of audio to image relationship formation and image to image relationship formation, with limited resources, and in real time.

\begin{figure}[h]
\begin{center}
\includegraphics[width=1.0\linewidth]{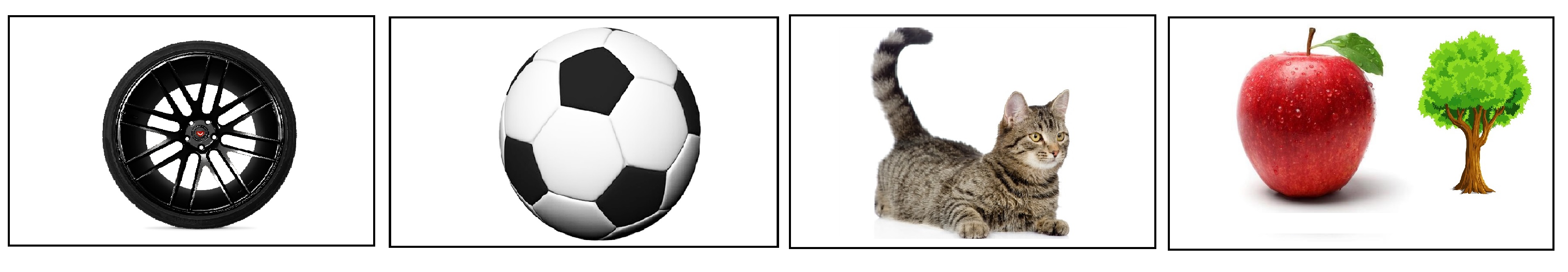}
\end{center}
\caption{Relationship Training}
\label{fig:reltraining}
\end{figure}

The program was started with an empty knowledge base. It was presented a few pictures as shown in figure \ref{fig:reltraining}, one after the other, by holding steady in front of the computer camera, each picture for between one and two minutes. During this time, an interacting user repeatedly spoke the word corresponding to the picture being shown, namely, “Wheel”, “Football”, “Cat” and “Apple”. After this brief “training”: (a) a blank picture was presented to the program, and the user spoke the four words spoken earlier, a few times each (b) a picture of just the tree, which was earlier part of the apple picture, was shown to the program for 10-15 seconds. For the two post-training steps, the future-trees predicted by the program were inspected.

\begin{figure}[h]
\begin{center}
\includegraphics[width=0.8\linewidth]{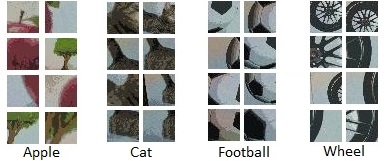}
\end{center}
\caption{Observations: Image recall for words}
\label{fig:imagerecall}
\end{figure}

\textit{Observations from the post-training future-tree inspection for (a):}\\
Figure \ref{fig:imagerecall} shows the images that occurred with high probabilities in the future-trees of the program, immediately after the corresponding words were spoken post-training. Images that were too similar to the ones shown below have been omitted. Looking at these, we can say that the program was able to recall images associated with the words with a high accuracy. The images recalled also had the spatial relationships amongst themselves through the image-focus-change action’s x,y,zoom pixel differences. 

There were many instances, both during and after the training, when silence was captured as an audio node. This node got associated with parts of all the pictures shown to the program, resulting in very low correlation with any specific image. The prediction functionality can discard such low probability associations, and thus dissociating very common audio and image nodes from other nodes.

There were instances when the spoken words were captured as two subsequent audio nodes due to time offsets. This again does not cause issues in retrieval. If the nodes become too short and common, they get automatically dissociated from other nodes, otherwise they continue to remain instrumental in prediction. For example, there were instances during the training when the word “football” was captured by the program in two different audio nodes, the first node containing the sound “foot” and the second node containing the sound “ball”. As a result of this, when it captured the sound “foot” after the training, it was able to recall both – the audio node “ball” and the images associated with football, like below: \\
\begin{footnotesize}
\verb+AUD.FOOT->AUD.BALL->ATT.IMG->IMG.FOOTBALL1->IFA.80,50,0->IMG.FOOTBALL2+\\
\end{footnotesize}

\textit{Observations from the post-training future-tree inspection for (b):}\\
The predicted future-trees contained the images of the fruit apple among the images with highest probabilities. The audio node representing the word “Apple” was among the audio nodes with highest probabilities, the other audio nodes standing mostly for “A”, “PPle”, “App”, “Le” and silence. The images recalled also carried the spatial relationships amongst themselves through the image-focus-change action. Looking at the results, we can say that the program is able to recall correctly the sounds associated with an image and also the images associated spatially with an image.

\subsubsection{Experiment: Action learning}

The objective behind this experiment was to establish that variations in pleasure pain \( \Delta P\textsubscript{net}\) over the temporal observation paths can indeed be the basis of learning of actions.

The program was configured to speak only these eight words: Apple, Ball, Car, Cat, Hello, Mumma, Papa, Wheel, one at a time. No other constraint was set on speaking, which means it can speak the words in any order any number of times. The image focus was kept static so that the program would only capture the whole picture shown to it and not its parts. 

The program was started with an empty knowledge base. Then the picture of a wheel, the second one in figure \ref{fig:reltraining}, was presented to the program by holding it steady in front of the camera. During this time, positive and negative feedback were provided to the program according to the words it spoke, that is, if it spoke “wheel”, the comfort level was increased, when it spoke something else, the comfort level was either kept same or decreased. In two to three minutes of time, the program reached a stage, where it started repeatedly only speaking the word wheel and no other words. The same process was repeated with the picture of a football and then of a cat and providing feedbacks until it converged to the words football and cat respectively. 

After this “training”, the program was again shown the pictures of the wheel, the football and the cat. This caused the program to speak “wheel, “football” and “cat” repeatedly as per the picture shown. From these observations, we can conclude that the program can indeed be conditioned to speak certain words for certain pictures

A relevant extract from the program log is shown below where IMG.158 represents the picture of the wheel and SPK.w-i-l, SPK.k-A-r, SPK.k-{-t, SPK.p-A-p-A represent speech actions for words wheel, car, cat, papa respectively. The numbers in brackets represent the probability and expected pleasure-pain change in the specific connection of the future-tree. By way of the below future-tree, the program knows that upon seeing the picture of a wheel, the best thing to do is to speak the word wheel.
\begin{footnotesize}\begin{verbatim}
...
-->IMG.158--[0.00,-3.10]-->SPK.k-A-r---[1.00,-3.10]--> ...
-->IMG.158--[0.01,-3.10]-->SPK.k-{-t---[1.00,-3.10]--> ...
-->IMG.158--[0.00,-2.10]-->SPK.p-A-p-A---[1.00,-2.10]--> ...
-->IMG.158--[0.03,0.21]-->SPK.w-i-l---[0.57,0.22]--> ...
Best parent-child combo: IMG.158 , SPK.w-i-l-
Taking learned action : SPK.w-i-l-           
\end{verbatim}\end{footnotesize}

The experiments show that using the principles of raw learnings discussed earlier, it is indeed possible to establish spatial image-image, and image-audio relationships, to recall the relationships upon repeat observations and to condition the agent to take certain actions. The approach is also very resource efficient, as the experiments were conducted in real time on an extremely modest hardware and with very few training examples. Hardware details mentioned in appendix \ref{agentprogram}

\section{Generalized Learning}

Raw learning forms the foundation of all learning; however, it can only help in predicting situations when the situation is repeating with either the exact same inputs or very similar ones. Without being able to generalize and apply the past learnings to new scenarios, the agent would be very limited in intelligence. 

In the proposed model, the agent regularly goes into an offline or sleep state to generalize the raw observations. Generalization makes the knowledge base more generic and robust to matching. It includes the formation of interpolated sensory nodes for relaxed matching constraints, the formation of sensory properties nodes for specific matching and superimposition, and the formation of group nodes for simpler logic pathways. The agent does all of this mainly by finding out similar observation paths and then reducing them into a common path. Reduction of information in a careful way not only helps the agent learn rules about the environment and about itself, but it also eases the resource requirement of maintaining a large number of observation paths. A couple of approaches that can be used to reduce the observation paths are discussed below. Our agent program implements these to some extent. The agent program goes into the offline or generalization mode when the observation trace length exceeds a fixed threshold. In this state, the program stops processing regular sensory inputs and actions. 

\subsection{Merging of nodes}

In this approach, the agent goes through all the observation trees recently touched and tries to identify similar path pairs. It considers two paths as similar if the number of memory nodes that are dissimilar is small compared to the length of the path. Once similar path pairs are identified, it tries to identify memory node pairs across the two paths that are corresponding and different. It attempts to merge such node pairs into a single memory node, which, if successfully created, is used to create a common observation path. Suppose there are two similar observation paths as below:\\
\begin{footnotesize}
\verb+IMG.1 --> IMG.11 --> IMG.2 --> IMG.3 --> IMG.4+\\
\verb+IMG.1 --> IMG.12 --> IMG.2 --> IMG.3 --> IMG.4+\\
\end{footnotesize}
Here IMG.11 and IMG.12 represent the memory node pair that is corresponding but different in the two paths. The merge process would try to create a merged node, say IMG.M.13 by combining the two. Now the two observation paths can be reduced to one:\\
\begin{footnotesize}
\verb+IMG.1 --> IMG.M.13 --> IMG.2 --> IMG.3 --> IMG.4+\\
\end{footnotesize}
A merged node is like a regular node but with a mask that allows for some deviations. In merged image nodes, in addition to the pixel-wise HSL values, there are pixel-wise HSL mask values to indicate the deviation that is allowed between the HSL values of corresponding pixels of two images to still consider the two pixels as matching. 

\subsubsection{Image merging}

\begin{figure}[h]
\begin{center}
\includegraphics[width=1.0\linewidth]{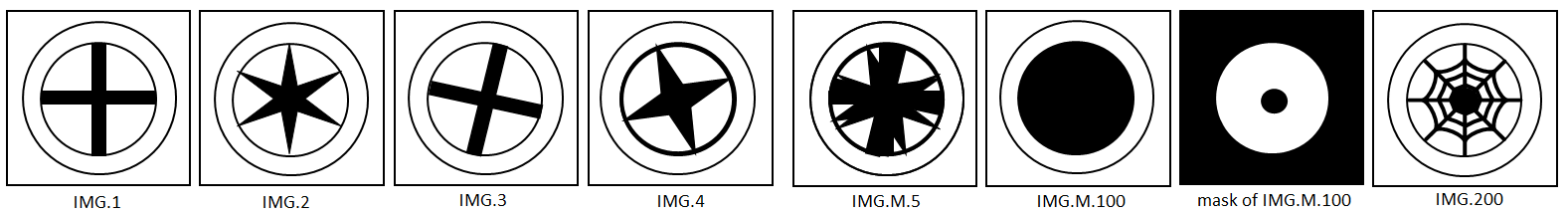}
\end{center}
\caption{Merging and masking}
\label{fig:wheelmerge}
\end{figure}

Suppose an agent is shown four different images of wheel and the word “wheel” is spoken before showing each of the images. Suppose the utterance of word wheel is represented by a single audio node AUD.WHEEL. The observation tree of AUD.WHEEL, or the four observation paths, would look something like:\\
\begin{footnotesize}
\verb+AUD.WHEEL --> ATT.IMG --|--(0.25)--> IMG.1 --> ...+\\
\verb+                        |--(0.25)--> IMG.2 --> ...+\\
\verb+                        |--(0.25)--> IMG.3 --> ...+\\
\verb+                        |--(0.25)--> IMG.4 --> ...+\\
\end{footnotesize}
If the agent decides to reduce these path segments by merging the above image nodes, the resulting merged image would look like IMG.M.5. Further observations and mergers may cause it to start looking like IMG.M.100 with a mask as depicted in the figure below. The black areas of the mask depict the portions of the candidate image to be matched with corresponding portions of IMG.M.100 to be considered a match of IMG.M.100. This means that the image IMG.200 which has never been seen before would still be considered a match for wheel. The resulting observation path after merger would be:\\
\begin{footnotesize}
\verb+AUD.WHEEL --> ATT.IMG --> IMG.M.100 --> ...+\\
\end{footnotesize}

In the same way, various properties of images can be learned, like color and pattern. Observations of images of Red apple, Red car, Red wheel, Red cap and so on along with the spoken word Red would cause the audio node for Red to get associated with an amorphous merged image of color Red. Similarly, the observations of Striped shirt, Striped pants, Striped bag and so on would cause an amorphous merged image with Striped pattern to get associated with the word Striped. The observation paths \\
\begin{footnotesize}
\verb+AUD.STRIPED -|--> AUD.SHIRT --> ATT.IMG --> IMG.1 --> ...+\\
\verb+             |--> AUD.PANTS --> ATT.IMG --> IMG.2 --> ...+\\
\verb+             |--> AUD.BAG   --> ATT.IMG --> IMG.3 --> ...+\\
\end{footnotesize} after merger of IMG.1, IMG.2 and IMG.3 into IMG.M.11 would become\\
\begin{footnotesize}
\verb+AUD.STRIPED --> JMP.1 --> ATT.IMG --> IMG.M.11 --> ...+\\
\end{footnotesize}
\begin{figure}[h]
\begin{center}
\includegraphics[width=0.65\linewidth]{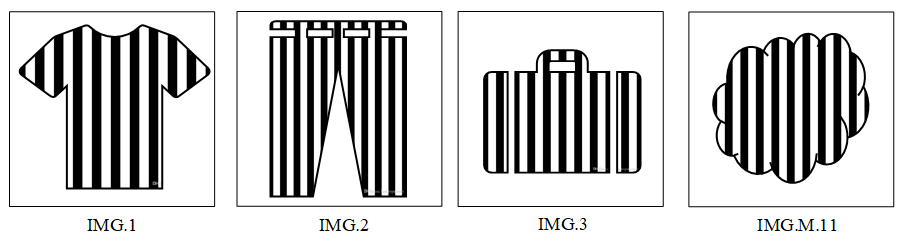}
\end{center}
\caption{Merging extracts image property (Striped)}
\label{fig:stripedmerge}
\end{figure}

If the images are too different for merging, an alternative way of how generalization can associate the "Striped" word sound with the striped pattern is shown in appendix \ref{stripezoom}

In addition to learning image properties, merging helps the agent learn images better by averaging noise and background differences over time.

\subsubsection{Audio merging}

The reduction and merging approach helps to generalize audio nodes as well. Suppose, the agent sees the image of an apple and hears the word Apple said by a male. Later it sees the same image of the apple and hears the word Apple said by a female. The observation paths would be:\\
\begin{footnotesize}
\verb+IMG.APPLE -> ATT.AUD -|--> AUD.APPLE_MALEVOICE   -> ...+\\
\verb+                      |--> AUD.APPLE_FEMALEVOICE -> ...+\\
\end{footnotesize}
If the agent decides to reduce the two observation segments, it would merge the two audio nodes and create a common audio node for the word apple with a mask that allows for some voice deviations\\
\begin{footnotesize}
\verb+IMG.APPLE --> ATT.AUD --> AUD.APPLE_GENERICVOICE -> ...+\\
\end{footnotesize}

In the same way, various properties of audio can be learned, like timbre, volume and pitch. Observations of various loud sounds would get associated with the word Loud and would eventually merge to create an amorphous merged audio node with high volume. Similarly, observations of many male and female voices would create gender specific timbre nodes.
\begin{footnotesize}
\verb+IMG.MALE_SPEAKER -> ATT.AUD -|--> AUD.APPLE_MALEVOICE1+\\
\verb+                             |--> AUD.CAR_MALEVOICE1+\\
\verb+                             |--> AUD.APPLE_MALEVOICE2+\\
\verb+                             |--> AUD.CAR_MALEVOICE3+\\
\verb+                             |--> AUD.WHEEL_MALEVOICE4+\\
\end{footnotesize}
would merge to become\\
\begin{footnotesize}
\verb+IMG.MALE_SPEAKER -> ATT.AUD -> AUD.MALEVOICE_TIMBRE+
\end{footnotesize}

It is possible that a similar audio merging happens inside a human baby's mind. Werker’s work \citep{werker1984cross} has shown that infants can distinguish between similar sounds of different languages, but later lose this ability.

The generalization process would similarly try to merge taste nodes, touch nodes, action nodes, jump nodes and so on. The details of the merging and mask generation however would depend upon the node-type. As we have seen in some examples, some of the merged nodes would start losing a recognizable form and thus would become more amodal than modal symbols.

\subsection{Grouping of nodes}

In this approach too, the agent first tries to identify similar path pairs and then the memory node or segment pairs that are corresponding and different across the path pair. Here the two corresponding nodes are either too different to be merged, or are of different types altogether, so instead of merging them into one, the agent creates a Group node by clustering the nodes or path segments together and thus enabling reduction of the two observation paths into one. 

Suppose the agent listens to three sentences: (i) A boy named John went to a park (ii) A boy name Andy went to a park, and (iii) A boy named Will went to a park. The observation paths would look like:\\
\begin{footnotesize}
\verb+AUD.A_BOY -> AUD.NAMED -|-> AUD.JOHN -> AUD.WENT -> AUD.TO_A -> AUD.PARK+\\
\verb+                        |-> AUD.ANDY -> AUD.WENT -> AUD.TO_A -> AUD.PARK+\\
\verb+                        |-> AUD.WILL -> AUD.WENT -> AUD.TO_A -> AUD.PARK+\\
\end{footnotesize}
The generalization process would create a group node, say GRP.NAMES by grouping audio nodes for "John", "Andy" and "Will". Now the observation paths can be reduced to one like :\\
\begin{footnotesize}
\verb+AUD.A_BOY -> AUD.NAMED -> GRP.NAMES -> AUD.WENT -> AUD.TO_A -> AUD.PARK+\\
\end{footnotesize}

To predict on the basis of an observation path is like running IF-AND-THEN conditions; for example, in the above scenario, If audio "a boy" is heard followed by audio "named" followed by audio "john", then the next nodes to expect are audio "went", "to a", "park". In that sense, a group node creates an OR condition on top of the inherent if-and-then logic. The grouping process doesn’t itself create any interpolations but helps create reduced paths that can now more easily participate in the processes that create interpolations or extrapolations. Also, a highly varied group node can become a Jump node; a jump node is like a wildcard node to which any node of any type would be considered a match while determining the conformation to an ongoing prediction. The group nodes can also be made global by the agent to be used in other observation paths where they don't exist yet but one or more of their constituent nodes do. 

\subsection{Further ways of generalization}

The discussed methods of generalization are perhaps a small subset of the ways the human brain is generalizing. By employing more mechanisms, like merging successive nodes within a single path, or by merging an unequal number of nodes across two paths or by merging observation paths where the indexing nodes are altogether different, and so on, the agent should be able to develop more effective generalizations and rules.

\subsection{Hypothesis evaluation}

We evaluate the generalization hypothesis through a simple experiment on the RTOP agent program. The purpose of the experiment is to show that by applying reduction and merging on the raw observation trees, interpolated images with mask can be formed. 

The program was configured to change the image focus only horizontally and not vertically. Also, the focus size was kept fixed. The program was then started with an empty knowledge base. First it was presented with a picture of an apple with a tree with some bush to the left of the tree. The picture was kept steady before the computer camera between one and two minutes. After that the program was presented with a picture of an apple with a tree with some bush to the right of the tree. Again, the picture was kept steady before the camera between one and two minutes. 

\begin{figure}[h]
\begin{center}
\includegraphics[width=0.6\linewidth]{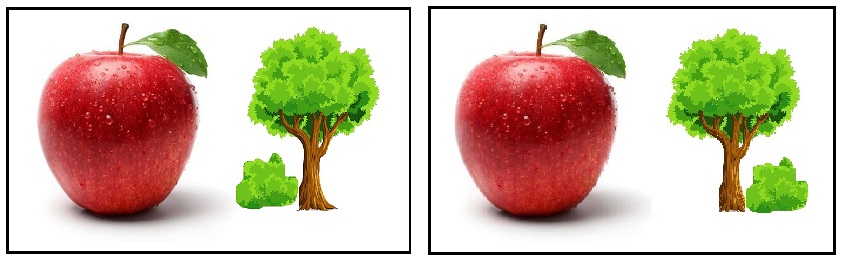}
\end{center}
\caption{Training for merger}
\label{fig:mergetraining}
\end{figure}

The subsequent generalization process of the program resulted in the creation of a merged node (the third of the below images). This merged node allows for bush to be either to the left or to the right of the tree or to be on both sides of the tree, while matching a new image to it and while evaluating if an image conforms to it in the ongoing prediction

\begin{figure}[h]
\begin{center}
\includegraphics[width=0.6\linewidth]{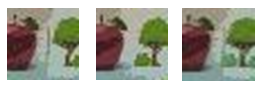}
\end{center}
\caption{Merging of image nodes}
\label{fig:mergeresults}
\end{figure}

A relevant extract of the program log is shown below. Here IMG.9 and IMG.218 represent images of the apple alone and the tree alone respectively. IMG.11 and IMG.191 represent the first two of the above three images. IFA.288 represents moving eye focus right by 288 pixels while IFA.155 represents moving eye focus right by 155 pixels
\begin{footnotesize}\begin{verbatim}
...
Attempting reduction of path-pair: 
 IMG.9->IFA.288,0,0->IMG.191->IMG.191->IFA.155,0,0->IMG.218
 IMG.9->IFA.288,0,0->IMG.11 ->IMG.11 ->IFA.155,0,0->IMG.218
Attempting merging of IMG.191 and IMG.11 
Created merged node IMG.M.1305 by merging IMG.191 and IMG.11
...
\end{verbatim}\end{footnotesize}

\section{Innovative Learning}

Raw learning and generalization can help the agent develop a representation of objects, their properties and their relationships, however, developing a model of the relationships themselves or of the transformations of objects (like most prepositions, adjectives, verbs or adverbs) and be able to apply these relationships or transformations to arbitrary objects, require forming indirect and innovative relationships among observed nodes. Raw and generalized learning are not sufficient to create internal projections that are too different from the ones encountered before. In the absence of such projections, agent would not be able to understand sentences or stories or solve problems. To be able to develop such concepts, the agent needs another inherent ability called Superimposition. We discuss below how superimposition should work, but this is yet to be implemented in the agent program for the validation purpose. A prerequisite to good innovative learning would be strong generalization both within and across indexing nodes.

\subsection{Parameterized paths and Superimpose action}

Suppose an agent has a well-developed raw learning. Suppose it now hears a phrase of words it has not heard before. Each word of the phrase would give rise to a different prediction. Each of these predictions may contain relationships to different images. The model proposes that the agent would try to reduce the number of predictions by merging some of them together. It would achieve the merger of predictions by superimposing the image and audio nodes across the predictions wherever possible. Reducing the predictions would help the agent conserve resources for managing the predictions, perhaps it is the same reason why the agent does merging of observations for generalization.

Superimposition is a special mental action operating on top of predicted image, audio and action nodes. This action requires the agent’s attention. Since the novel images or audios generated through superimposition are produced under the agent’s attention, they get stored in the memory. Listening to a story would generate many novel images created through superimposition of images corresponding to the words used in the story. It is these images that help the agent to remember the story and be able to recall it quickly.

The development or learning of this superimpose action would happen the same way regular actions are learned. In the initial stages, the agent would randomly superimpose images or audios across different predictions. Through the process of reward and penalty, it would learn which images or audios to pick from which nodes of the different predictions for superimposition. Correct predictions may well be rewards themselves. This process would also gives rise to observation paths that have nodes that always require a superimposition. Such observation paths we term as Parameterized observation paths. Below are presented a couple of scenarios to show how such abilities can give rise to higher level understanding in the agent.

\subsection{Superimposition of Images}

We have seen in the generalized learning section how the agent would develop a merged image corresponding to the sound "Striped". By using superimposition action over time, the agent would learn that this merged striped image IMG.STRIPED needs to be superimposed on the most predicted image (PRED\_IMG) whenever it is pulled into a prediction. Suppose the agent now hears the phrase “Striped Apple” for the first time. The future trees would be:\\
\begin{footnotesize}
\verb+AUD.STRIPED -> ATT.IMG -> SIA.[PRED_IMG,IMG.STRIPED] -> ... +\\
\verb+AUD.APPLE -> ATT.IMG -> IMG.APPLE -> ...+\\
\end{footnotesize}
where SIA stands for superimpose action. The most predicted image being IMG.APPLE, the projection would become\\
\begin{footnotesize}
\verb+-> SIA.[IMG.APPLE,IMG.STRIPED] -> ...+
\end{footnotesize}

\begin{figure}[h]
\begin{center}
\includegraphics[width=0.6\linewidth]{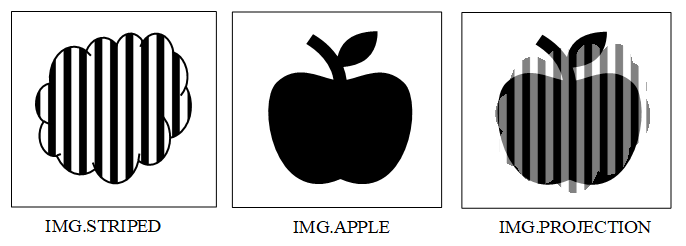}
\end{center}
\caption{Superimposition of image nodes}
\label{fig:supstripedapple}
\end{figure}

Adjectives may be easier to extract and superimpose; let’s take a look at how verb superimposition might work. Suppose the agent hears the phrases “apple falls” and “cap falls” followed by their corresponding images as shown in figure \ref{fig:fallsraw}. In each case, first the object is seen at a height, then is seen a hazy image of the object moving down, then the image focus moves down, and finally the object is seen on the floor. Assuming each spoken word is one audio node, the observation paths are:\\
\begin{footnotesize}
\verb+AUD.APPLE -> AUD.FALLS -> ATT.IMG -> IMG.1 -> IMG.2 -> IFA.DOWN -> IMG.3+\\
\verb+AUD.CAP -> AUD.FALLS -> ATT.IMG -> IMG.11 -> IMG.2 -> IFA.DOWN -> IMG.13+\\
\end{footnotesize}
\begin{figure}[h]
\begin{center}
\includegraphics[width=1.0\linewidth]{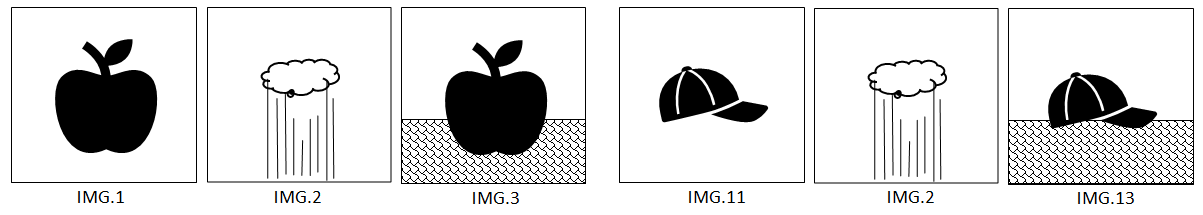}
\end{center}
\caption{Falls - Raw}
\label{fig:fallsraw}
\end{figure}

If there are many similar observations for different item drops, then a desirable generalization algorithm would reduce paths and merge nodes as shown in figure \ref{fig:fallssup}. Subsequent superimposition learning and parametrization would result in:\\
\begin{footnotesize}
\verb+AUD.FALLS->ATT.IMG->SIA.[P_IMG,IMG.M.101]->IMG.2->IFA.DOWN->SIA:[P_IMG,IMG.M.103]+\\
\end{footnotesize}
In other words, prediction for node AUD.FALLS would become a function that takes as parameter a predicted image (P\_IMG), that is, the most likely image from ongoing predictions and applies the superimpose action (SIA) by placing P\_IMG onto IMG.M.101 and IMG.M.103 on the projection. Suppose the agent now hears the phrase “pen falls”, a phrase it has never heard before. Prediction for AUD.PEN would be:  \begin{footnotesize}
\verb+AUD.PEN -> ATT.IMG -> IMG.PEN +
\end{footnotesize}
with the most predicted image (P\_IMG) being IMG.PEN. Applying this as a parameter in the prediction for AUD.FALLS provides the below projection: \\
\begin{footnotesize}
\verb+SIA:[IMG.PEN,IMG.M.101] -> IMG.2 -> IFA.DOWN -> SIA:[IMG.PEN,IMG.M.103]+
\end{footnotesize}

\begin{figure}[h]
\begin{center}
\includegraphics[width=1.0\linewidth]{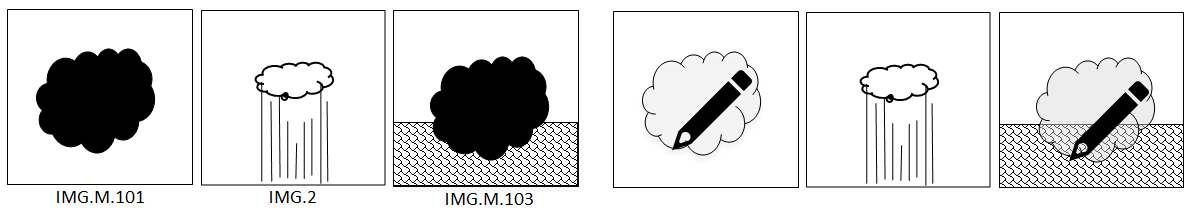}
\end{center}
\caption{Falls - Merged and Superimposed}
\label{fig:fallssup}
\end{figure}

\subsection{Superimposition of Audio}

Superimposition would apply to audio nodes the same way as it does to image nodes. A human brain can imagine a sentence being said in a variety of ways – in male voice, in female voice, spoken loudly, spoken soft, spoken in anger, spoken in love and so on. The agent can achieve a similar functionality through superimposition of audio nodes.

We have seen in the generalization section how merging would give rise to audio nodes for words that are neutral or generic in terms of voice quality. The same generalization process would give rise to audio nodes for male voice timbre and female voice timbre. Suppose an agent where such nodes are already created is told that “The woman said ‘car’”. By using superimposition action over time, it would learn that “said” corresponds to superimposition of two audio nodes P\_PRECEDING and P\_FOLLOWING. \\
\begin{footnotesize}
\verb+AUD.WOMAN -> AUD.WOMAN_VOICE_TIMBRE+\\
\verb+AUD.SAID -> SIA.[P_PRECEDING, P_FOLLOWING]+ \\
\end{footnotesize}
For AUD.SAID, the preceding audio being AUD.WOMAN\_VOICE\_TIMBRE and the following audio being AUD.CAR\_GENERIC, the projection would be\\
\begin{footnotesize}
\verb+-> SIA.[AUD.CAR_GENERIC, AUD.WOMAN_VOICE_TIMBRE] +\\
\end{footnotesize}

\subsection{Thought}

We have seen how the agent can merge memory nodes across different predictions for different spoken words to create a common projection. When the number of predictions become too many, then achieving a common reduced projection may involve multiple steps of superimposition. It may also require iteratively superimposing and determining which merger makes the most sense, that is, is closest to some past or current observation or determining which one is the most beneficial. Alternatively, if the number of predictions is too low, then to populate the parameterized paths may require scanning the knowledge base to find nodes that fit the paths. These are resource intensive processes and would require agent to extend attention to this superimposition process for a longer duration. In such situations, the agent can be said to be involved in thought. 

The superimpositions created by thought get stored to memory. Sometimes, the superimposition may have to be left incomplete, that is, with unfulfilled parameters, in which case the incomplete projection would get stored to the memory. Ongoing thoughts may uncover prior incomplete projections which can further extend the agents involvement in thought. Background processing to capture audio and visual inputs and do quick and dirty predictions would still go on while the brain is busy in thought. A threat or opportunity event detected in the background processing can cause the brain to shift attention from thought to the event.

\section{Conclusion and future work}

We have seen how the three proposed mechanisms of intelligence, namely, raw learning, generalization and innovation can be built on top of the proposed generic observation tree structure, for the development of general intelligence using limited resources. The system built using this is approach is easy to understand, investigate and improve. While the agent program serves as a proof of selected concepts, it is far from the capabilities of a human brain. Several major enhancements would be required in the future to narrow this gap. The prediction algorithm would need several improvements in determining the current observation path as well as in looking it up in the knowledge base. The matching would perhaps need to be progressively made more specific with age by constricting thresholds as the agent builds a knowledge base. Generalization, as discussed earlier, needs further mechanisms to be able to develop a broader set of memory pathways and rules. Innovative learning principles have not been implemented so far and remain to be verified and demonstrated. The training of the agent program was done by a human interacting with it, like a human interacts with an infant. This approach is however not scalable, and new simulated methods of automated training would need to be developed. 

\newpage
\subsubsection*{Acknowledgments}
We thank Ashish Kapoor, Sr Principal Research Manager, Microsoft for the motivation and guidance

\bibliography{brain}
\bibliographystyle{iclr2020_conference}

\newpage
\appendix
\section{Appendix}

\subsection{Details about RTOP agent program}\label{agentprogram}

The program implements many of the principles discussed in this paper. 

Images are captured from the computer camera every few hundred milliseconds. An image is stored as 32x32 8bit HSL (3 bits for H, 2 bits for S, 3 bits for L). Images are indexed on summary values like mean-of-lightness, variance-of-lightness and so on. After shortlisting candidates through the index, image matching is done by comparing lightness of each pixel and checking overall difference against a threshold.

Audio is captured from the computer microphone. Successive audio nodes are captured after a gap of few hundred milliseconds. Audio is stored as a single channel 8bit PCM waveform of 800ms duration with sampling rate of 16k/s. Candidate shortlisting on index and subsequent matching, both are done on summary values like variance-of-amplitude, mean-cross-rate of amplitude and so on. 

Hunger is simulated in the program; it increases by a scalar value after every few seconds. A single compound input has been created called comfort to represent the overall pleasure pain across different internal senses. This too takes one of several scalar values. An interacting user can reward the program by either providing a “feed” which quenches the hunger, or by increasing the comfort level. 

The actions built in the program are attention change, image focus change and speech action. The action generation happens in a probabilistic way, where the probabilities depend upon various parameters like time since the last action, repetition of input data, presence of prediction, etc. A high amount of change in the audio or visual area also causes change in attention. The unit of image focus change is presently pixels; it would be useful to have, in addition, represent the focus change as a percentage of the current focus area. Image focus change action does not actually interact with the camera, it simply crops the image captured by the camera as per the focus parameters and then passes the cropped image to visual input processor. Speech action is generated by sending randomly generated phones to the Mbrola software, which plays the sequence of phones by using pre-recorded sounds for a large number of phone combinations. The phone sequence generation is random but with probabilities of phone generation skewed in the favour of few consonants and few vowels. The duration of consonants is fixed at 100ms and that of vowels is fixed at 500ms. The sound-database and loudness and frequency are fixed for simplicity

The program is written in Java and uses Java JRE 1.8 32-bit

For the purpose of the experiments, the hardware used was HP Laptop 2.16 GHz Quad Core Processor with Hyper V and having 4GB RAM running 64-bit Windows 10 OS. Image is captured through the built in HP Truevision HD webcam while audio is captured through the built in Microphone Realtek High Definition Audio. 

\subsection{Generalization of striped property through zoom}\label{stripezoom}

\begin{figure}[h]
\begin{center}
\includegraphics[width=0.7\linewidth]{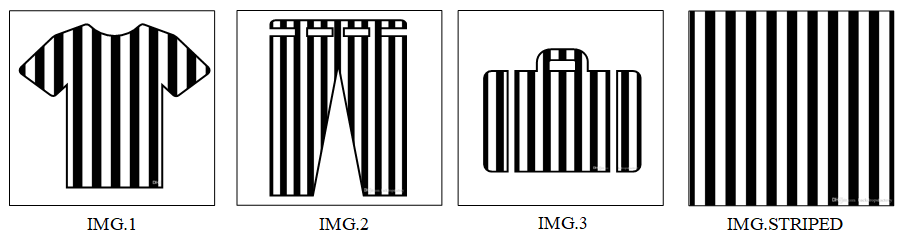}
\end{center}
\caption{Stripes}
\label{fig:stripezoom}
\end{figure}

The observations of Striped shirt, Striped pants, Striped bag upon zooming focus all show the same Striped pattern. The observation tree  would be: \\

\begin{footnotesize}
\verb+AUD.STRIPED -|--> AUD.SHIRT --> ATT.IMG --> IMG.1 --> IFA.ZOOM --> IMG.STRIPED +\\
\verb+             |--> AUD.PANTS --> ATT.IMG --> IMG.2 --> IFA.ZOOM --> IMG.STRIPED +\\
\verb+             |--> AUD.BAG   --> ATT.IMG --> IMG.3 --> IFA.ZOOM --> IMG.STRIPED +\\
\end{footnotesize} 
After shortening and merging, the observation can become\\
\begin{footnotesize}
\verb+AUD.STRIPED --> JMP.4 --> IMG.STRIPED +\\
\end{footnotesize}

\subsection{Architecture and Flow diagram of RTOP Agent}\label{rtoparch}

\begin{figure}[h]
\begin{center}
\includegraphics[width=1.1\linewidth]{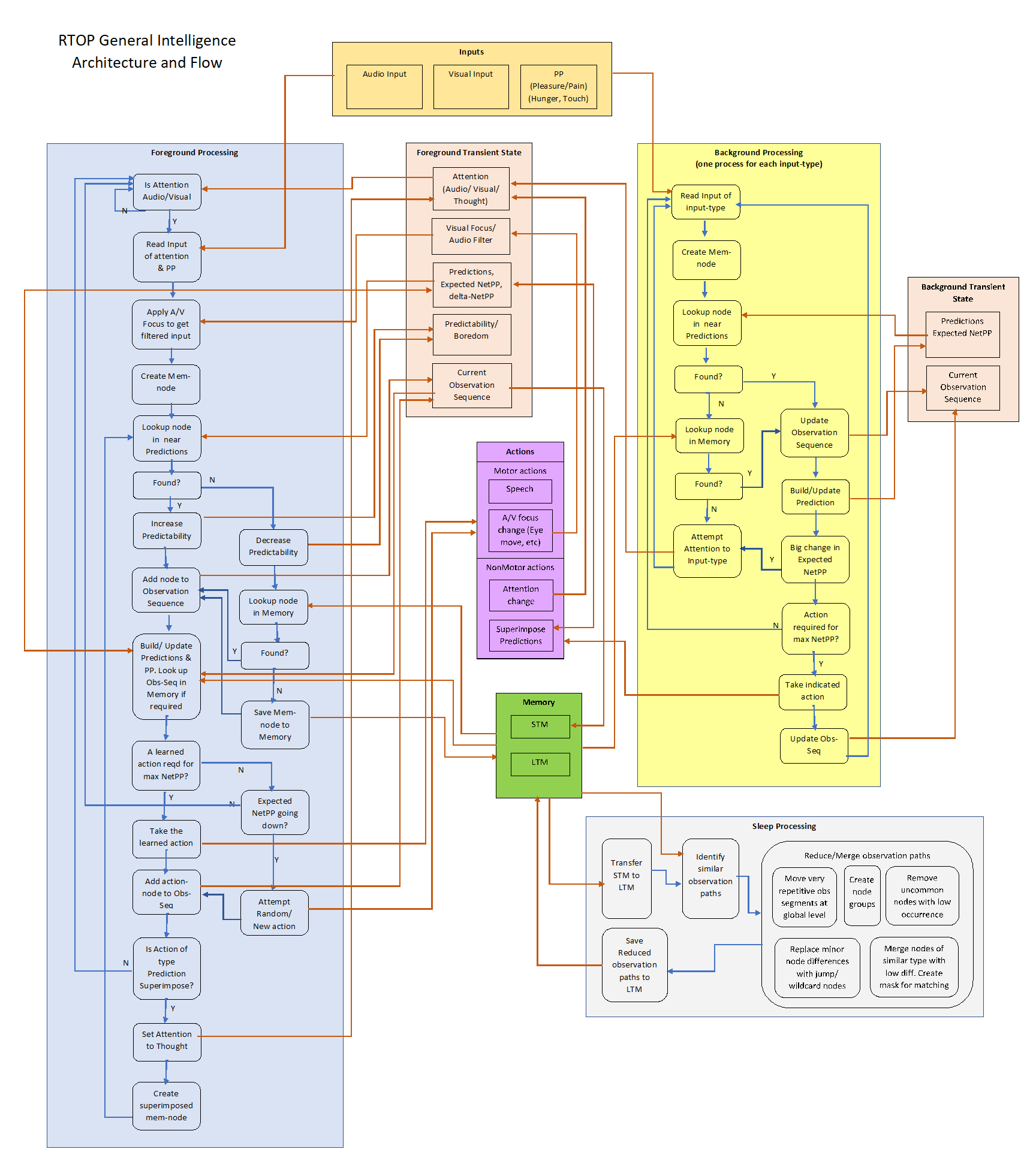}
\end{center}
\caption{Architecture and Flow}
\label{fig:rtoparch}
\end{figure}

\end{document}